\documentclass{article}
\usepackage{ijcai21}

\usepackage{times}
\usepackage{soul}
\usepackage{url}
\usepackage[hidelinks]{hyperref}
\usepackage[utf8]{inputenc}
\usepackage[small]{caption}
\usepackage{graphicx}
\usepackage{amsmath}
\usepackage{amsthm}
\usepackage{booktabs}
\usepackage{algorithm}
\usepackage{algorithmic}
\graphicspath{ {./imgs/} }

\title{Context-Dependent Semantic Parsing for Temporal Relation Extraction}
\author{
Bo-Ying Su$^1$
\and
Shang-Ling Hsu$^2$\and
Kuan-Yin Lai$^3$\And
Jane Yung-jen Hsu$^3$
\affiliations
$^1$University of California, San Diego\\
$^2$
The Hong Kong University of Science and Technology\\
$^3$National Taiwan University\\
\emails
b1su@ucsd.edu,
shsuaa@connect.ust.hk,
eddy50811@gmail.com,
yjhsu@csie.ntu.edu.tw
}
\begin{document}

\maketitle

\begin{abstract}
  Extracting temporal relations among events from unstructured text has extensive applications, such as temporal reasoning and question answering. While it is difficult, recent development of Neural-symbolic methods has shown promising results on solving similar tasks. Current temporal relation extraction methods usually suffer from limited expressivity and inconsistent relation inference. For example, in TimeML annotations, the concept of intersection is absent. Additionally, current methods do not guarantee the consistency among the predicted annotations. In this work, we propose SMARTER, a neural semantic parser, to extract temporal information in text effectively. SMARTER parses natural language to an executable logical form representation, based on a custom typed lambda calculus. In the training phase, dynamic programming on denotations (DPD) technique is used to provide weak supervision on logical forms. In the inference phase, SMARTER generates a temporal relation graph by executing the logical form. As a result, our neural semantic parser produces logical forms capturing the temporal information of text precisely. The accurate logical form representations of an event given the context ensure the correctness of the extracted relations.
\end{abstract}

\section{Introduction}

Extracting temporal relationships between events in unstructured text is one of the core tasks in Natural Language Understanding (NLU). Temporal Information Extraction (TIE) helps gain a holistic view on the context of a paragraph, which includes what, when and how long did a event happen. By knowing the temporal relations between events, TIE enables a machine to reason on the timelines of incidents, allowing applications such as Question Answering and Text Summarization.

TIE requires a high-level understanding in a language. Thus, is a difficult task to be resolved and is still an active research topic in NLP.  One of the challenges is that most temporal hints in the text, which seem to be obvious to humans, are implied in the language. TIE requires the algorithm to ``read between the lines". Consider the following example: ``The dog is running out the house to bite the man who stands in the yard.” Human can tell from the text that “man stands in the yard” happens before “dog runs out”, but for nature language processors, there is no explicit keywords, such as ``before" or ``after", that could be found in the sentence. Additionally, while humans recognize that ``running out" and ``bites" are events that happen instantly and ``stand" would continue for a period of time, machines don't. To capture subtle interactions between events like the one described above, we present SMARTER, a nature language model that does TIE by utilizing linguistic information and common sense knowledge.


We propose a neural semantic parser for temporal information extraction from text, named SMARTER (context-dependent SeMantic pARsing for TEmporal Relation extraction). SMARTER parses natural language to an executable logical form representation, based on a custom typed lambda calculus. In inference phase, the logical form is being executed to generate a temporal relation graph which is solved by a constraint solver. In training phase, dynamic programming on denotations (DPD) technique is used to provide weak supervision on logical forms. Our contributions are as follows:
\begin{itemize}
    \item We propose SMARTER, a neural semantic parser that captures compositional temporal information from text.
    \item We propose a grammar that is able to express compositional temporal information precisely.
\end{itemize}

This paper is structured as follows: Section 2 introduces the related work in temporal relation extraction and semantic parsing. Section 3 describes every component of our approach. Section 4 presents our experimental results. A conclusion is given in Section 5.


\section{Related Work}
This work adopts a context-dependent semantic parsing approach for the temporal relation extraction task.

Temporal relation extraction from text comprises several sub-tasks, such as the identification of event expressions and temporal relations between events and times, as formally defined in SemEval-2016 \cite{bethard2016semeval}. While previous studies on the former task, such as the ones based on support vector machines \cite{lee2016uthealth} and conditional random fields \cite{grouin2016limsi}, achieve near-human performance, similar methods applied to the latter task still present a gap between the state-of-the-art systems and human performance \cite{bethard2016semeval}. 
Fortunately, experiments with the convolutional neural network (CNNs) \cite{dligach2017neural} and long short-term memory neural networks (LSTMs) \cite{tourille2017neural} show that they outperformed hand-engineered feature-based models. Structured learning approaches are also proposed to enhance performance by incorporating more knowledge \cite{ning2017structured}.
Recent studies explore the power of pre-trained self-attentive models (Transformer) on temporal relation extraction by proposing extensions and variants \cite{wang2019extracting,ning2019improved,yang2019exploring},
but such methods lack the capability to express the compositional temporal relations among multiple events. To this end, we propose to address a problem in a neural semantic parsing approach, applying Transformer as the input embedding model.

Given a question, typical semantic parsing approaches for question answering in a knowledge base learn a model to transform a question to an executable logical form for answer retrieval. Early approaches are rule-based, mapping questions into database queries \cite{zelle1996learning,kwiatkowksi2010inducing}, which requires labor-intensive logical form annotations. To address the problem, later works train semantic parsers bypassing annotated logical forms with weak supervision \cite{berant2013semantic,liang2017neural}. 
Recent approaches further introduce the encoder-decoder neural network architecture in semantic parsing \cite{jia2016data,krishnamurthy2017neural}. Such an architecture is then adopted in AllenNLP \cite{gardner2018allennlp} as an open-source software for natural language processing, and the semantic parser in this work is inspired by and built upon the AllenNLP implementation.


\section{Our Approach}
\begin{figure}[t!]
    \centering
    \includegraphics[scale=0.7]{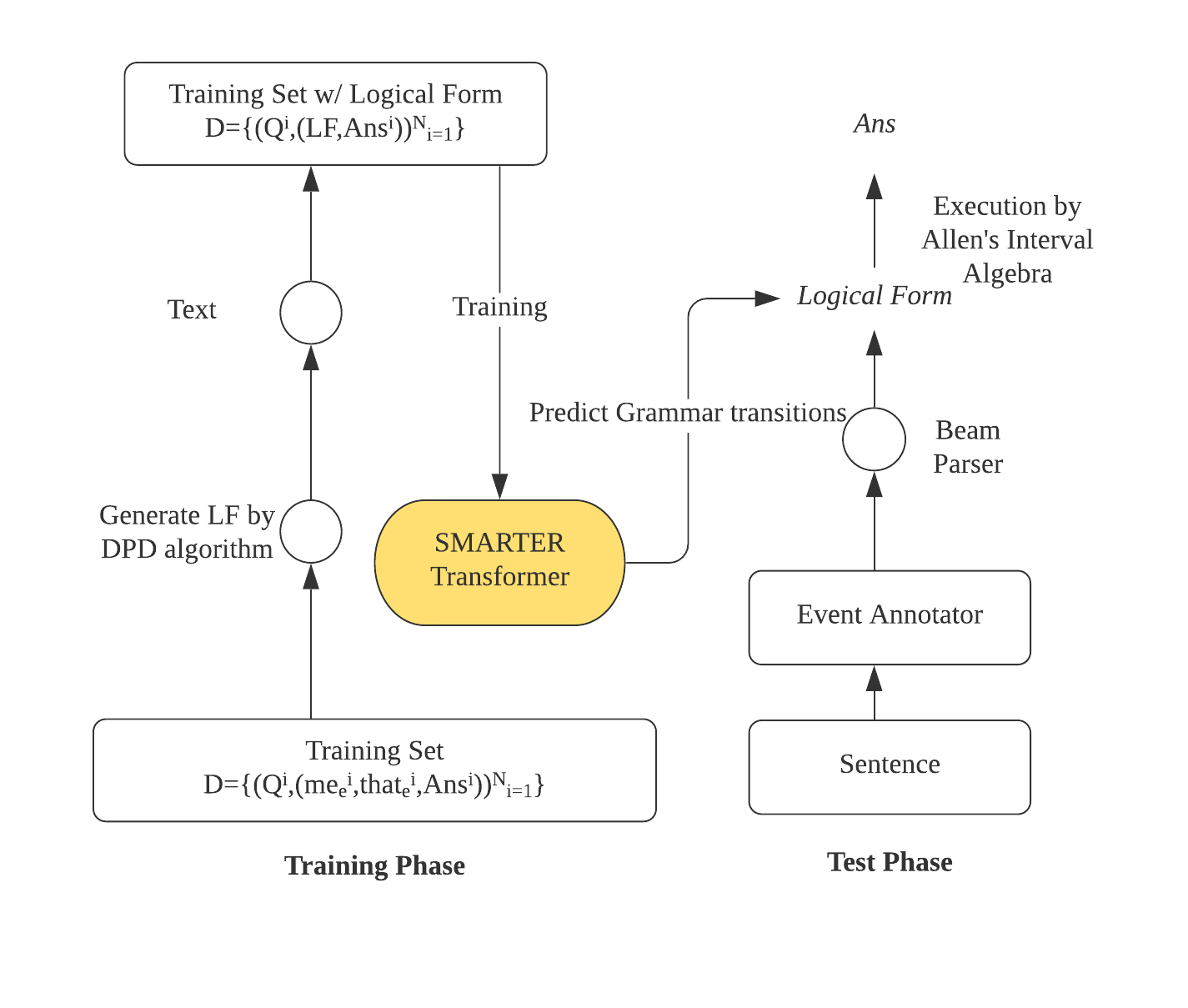}
    \caption{An overview of the proposed approach}
    \label{fig:overview_of_proposed_approach}
\end{figure}

Our approach to the Temporal Information Extraction task is mainly based on the neural semantic parsing architecture, adapted to the temporal reasoning domain shown in Figure \ref{fig:overview_of_proposed_approach}. Our novel neural semantic parsing is able to extract temporal relations between events and time expressions in a paragraph effectively, outperforming other previously proposed methods. 

SMARTER is composed of two parts, Event Annotator and Relation Extractor. The Event Annotator module is based on BERT with attention mechanisms. The Relation Extractor module is based on the Encoder-Decoder Transformer architecture that produces logical forms that can be executed to obtain the answer to the query. Our model is trained on multiple datasets that is annotated according to the TimeML \cite{pustejovsky-etal-2203-timeml} specification. 

\subsection{Interval Algebra}
  James F. Allen describes the thirteen basic relations between time intervals \cite{allen1983maintaining}. We base our model on the Allen's Interval Algebra because 1. It is exhaustive. The relationship between any pair of definite time intervals can be described by one of the thirteen relations. 2. It is distinctive. In other words, the relationship between any pair of definite time intervals can only be related by exactly one of the thirteen relations.
  \subsubsection{The 13 Basic Relations}
  These are the 13 basic relations between intervals proposed by James F. Allen. \\
  \begin{figure}[t!]
    \centering
    \includegraphics[scale=0.3]{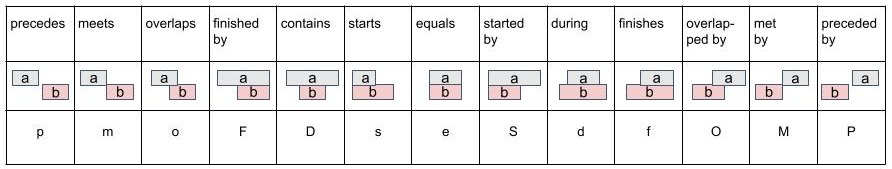}
    \caption{13 basic relations between intervals}
    \label{fig:13_basic_relations}
\end{figure}

  \subsubsection{The TimeML Dataset and Allen's Interval Algebra}
  The dataset we used to train our model follows the TimeML specification. TimeML annotation is largely inspired by Allen's Interval Algebra. The types of TLINKS that are annotated in the TimeML dataset have one to one correspondence to the 13 basic relations. 
  \subsubsection{The Network of Reference Intervals}
  Allen proposed a graph-based algorithm on temporal reasoning in his paper. The method involves creating Reference Intervals, which ``group together clusters of intervals for which the temporal constraints between each pair of intervals in the cluster is fully compute." The relationship between two intervals is then computed by searching up the reference hierarchy until all paths between the two nodes are found. Transitivity table is used to propagate the relation composition the in the process. Our model utilizes the   graph-based algorithm proposed by Allen to infer relations between events and time intervals.
  
  \subsubsection{Composition of Relations}
  When it comes relation composition, we refer to the transitivity table proposed by Allen's Interval Algebra.

\subsection{Logical Form}
  Logical forms are intermediate productions of our model that are eventually executed to obtain the answer (denotation) to the query. Instead of obtaining the answer directly, the decoder of our model generates logical forms. As almost all the temporal annotated datasets available contain only answers (denotations), we overcome the challenge of having weakly supervised training sets by generating logical forms and use them as a guidance of our model to obtaining the correct answer.
  
  The thirteen relations proposed by Allen's Interval Algebra forms the foundation grammar of SMARTER's logical form language. Every temporal relation of time interval can be represented as one of the predicates. We can view these relations as constraints on the time interval, which produces a reference time interval that encapsulates the original one. The set operations, including intersection and union, manipulate the intervals to capture the more comprehensive expressions of temporal information.

  \subsubsection{Language}
  The Language of SMARTER is based on simply typed lambda calculus, in the form of lisp tree. The predicates in the defined language consists of three classes of functions:
  1. Constants: The constant function returns a TimeInterval type that represents the interval of an event.
  2. Relation Functions: Relation functions create a reference interval from the input TimeInterval that returns a TimeInterval type. The Relation Functions have a one-to-one correspondence to the thirteen interval relationships defined in Allen's Interval Algebra.
  3. Set Operation Functions: Set operation functions are functions that accepts one or more TimeInterval as a input, and returns a TimeInterval type. Currently Union and Intersection operations are implemented in the language of SMARTER.
  By the nature of our language construct, any logical forms in SMARTER are guaranteed to be valid. That is, all logical forms are executable any will always product a result (although its denotation may or may not be correct).
  
  \subsubsection{Search for Logical Forms}
  In order to create a logical form training set for the weakly supervised training data, we have to search for correct logical forms that produce correct answers exhaustively. The method of DPD (Dynamic Programming on Denotations) training is employed in our data pre-processing process. The DPD training technique described in the original paper consists of an exhaustive search process, which iterates through all the valid productions in the defined language, then filter out incorrect logical forms by executing them and compare the results against the training set. The method is proven to be effective in finding correct logical forms; however, a large portion of the logical form productions are considered spurious. That is, some logical form generated is able to produce correct answers( denotations), but for the wrong reason. The correct denotations happen to be a coincidence because the logical form does not represent the real intentions of the question. We mitigate such an issue by preventing SMARTER's language to chain two functions whose composition is equivalent to any of the two. Our result show that a large amount of spurious logical forms are avoided by the measure and the action space is reduced at the same time, which speeds up the DPD training process.
  
  \subsubsection{Execute Logical Forms}
  The logical forms are executed by converting the logical form, which is a lisp tree, into Linearized Abstract Syntax Tree, called Action Sequence. The Action Sequence includes the transitions happen on the defined language using the Production Rules defined in the language specification. Then, by applying the Production Rules to the root TimeInterval object, the constraints between any events and reference intervals generated along the way is collected until the executor reaches the end of the Action Sequence. Finally, the constraints are converted to a network that composed of reference intervals and the graph search algorithm proposed by Allen's Interval Algebra is executed on the network to extract the relationship between all pairs of intervals.

\subsection{Preprocessor}
The input context is unstructured natural language. we need to first get the annotations of the events and time information.
\subsubsection{Event Annotator}
The event annotator is a Transformer with a hierarchical classification head on the top for token classification. Tokens are first classified into two categories: whether it is an event token. An event token is then further classified into another two categories: whether it is the last token of the event tokens.
\subsubsection{Time Annotator}
The time expressions are annotated as TimeX3 tags using SUTime \cite{chang2012sutime}.

\subsection{Model}
SMARTER follows the Transformer architecture. A sentence is being tokenized and embedded using pretrained BERT model. The sentence embedding is used as the initial hidden state of the LSTM cells in the decoder. The decoder decodes LSTM's state to guide the grammar transitions of the parser, generating the logical form representing the input query. \\
\begin{figure}[t!]
    \centering
    \includegraphics[scale=0.7]{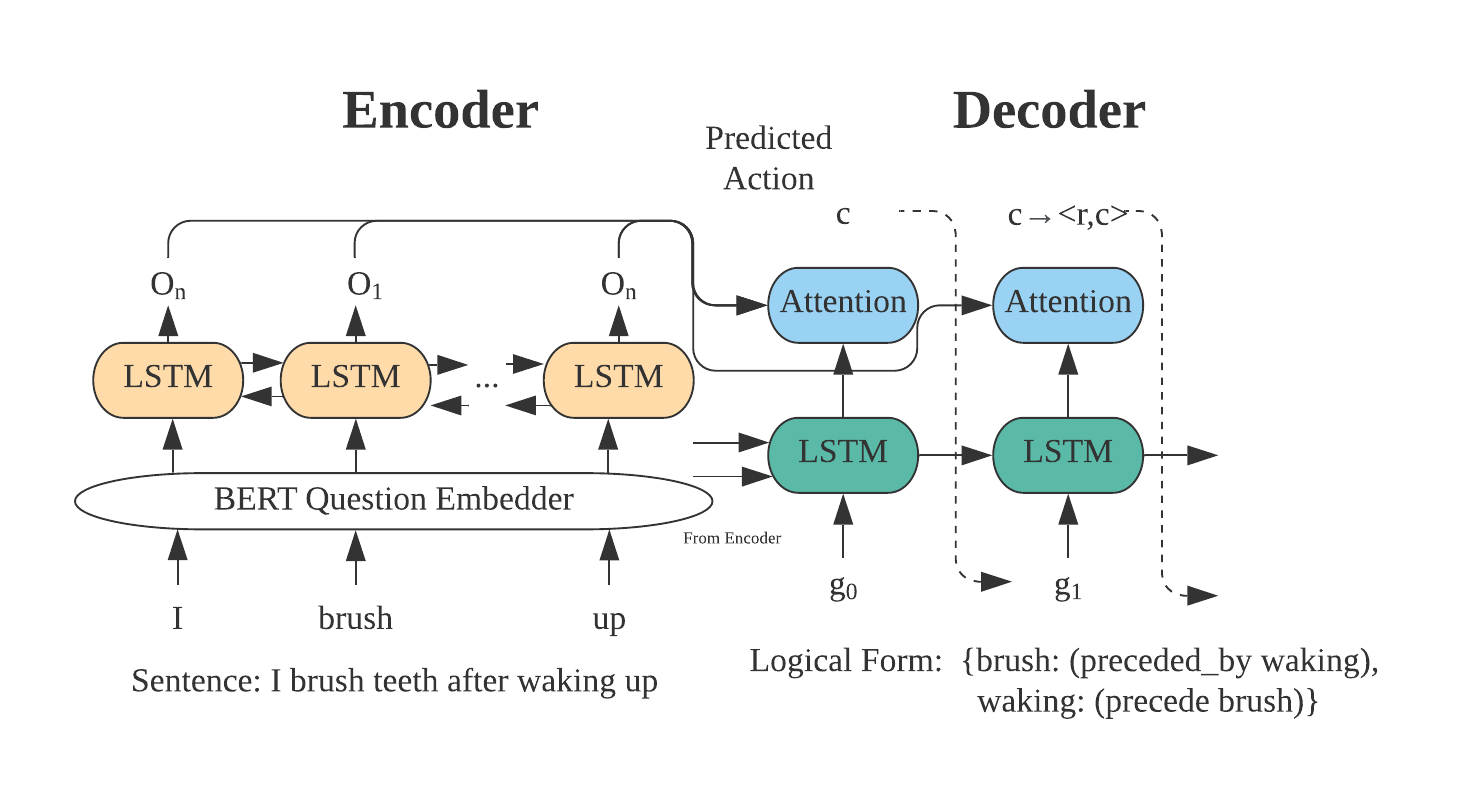}
    \caption{Transformer architecture of SMARTER}
    \label{fig:transformer_architecture_of_smarter}
\end{figure}

\subsubsection{Encoder}
\begin{figure}[t!]
    \centering
    \includegraphics[scale=1]{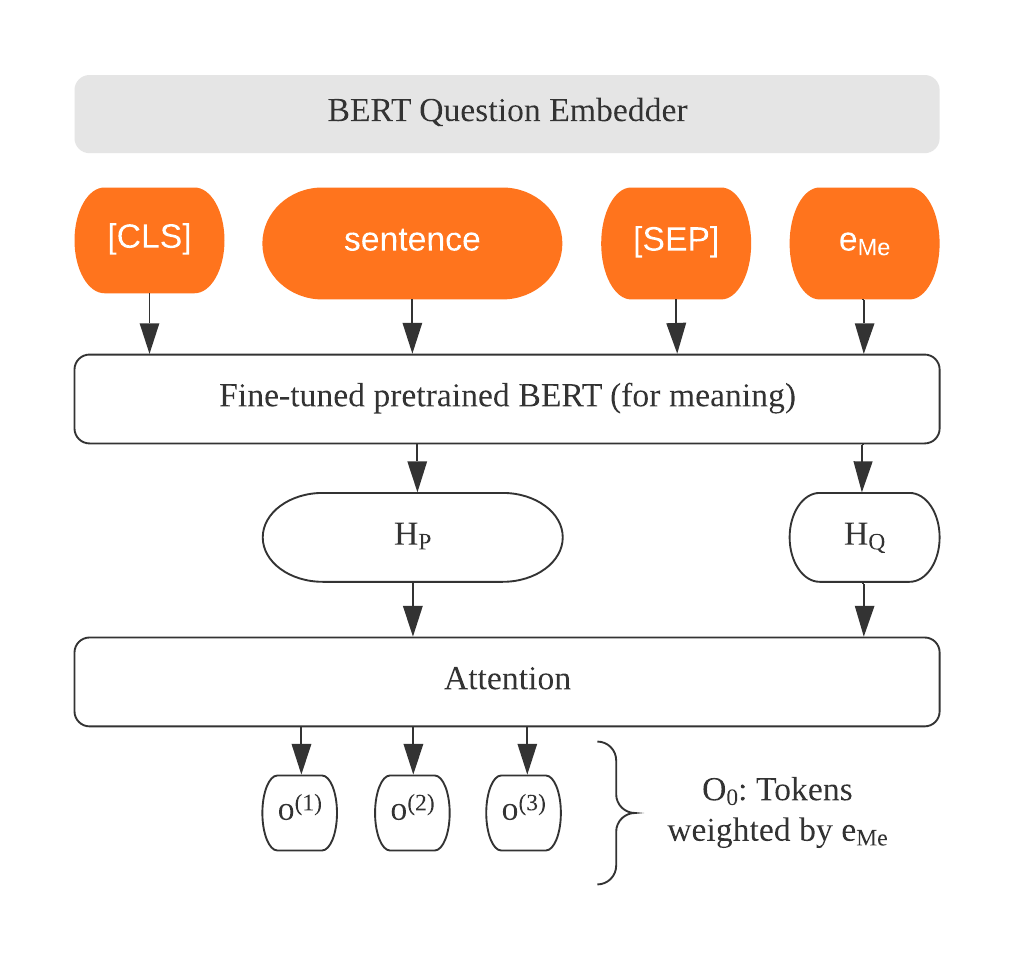}
    \caption{The architecture of BERT Question Embedder}
    \label{fig:arch_bert_q_embedder}
\end{figure}

Our encoder, as shown in Figure \ref{fig:arch_bert_q_embedder}, is a pretrained BERT model with attention mechanism, inspired by COSMOSQA \cite{huang2019cosmos}. The input sentence is first tokenized and concatenated with the target event tokens to make up a input sample. 
The input to the BERT model looks like this: [CLS] sentence\_tokens [SEP] me\_event\_tokens [SEP]

The BERT model projects the input sample to high dimensions and form the word embeddings, which contain the contextualized information. We utilize the sequential output of the bert model as the input of the downstream decoder component.

The attention mechanism here is used to provide weighted average on the imput sentence. The more ``important" a token is to the semantic parsing result, the more weight it gains. The attention weight captures the hidden relation and similarity of the temporal relation to the input sentence tokens. \\

\subsubsection{Decoder}
\begin{figure}[t!]
    \centering
    \includegraphics[scale=1]{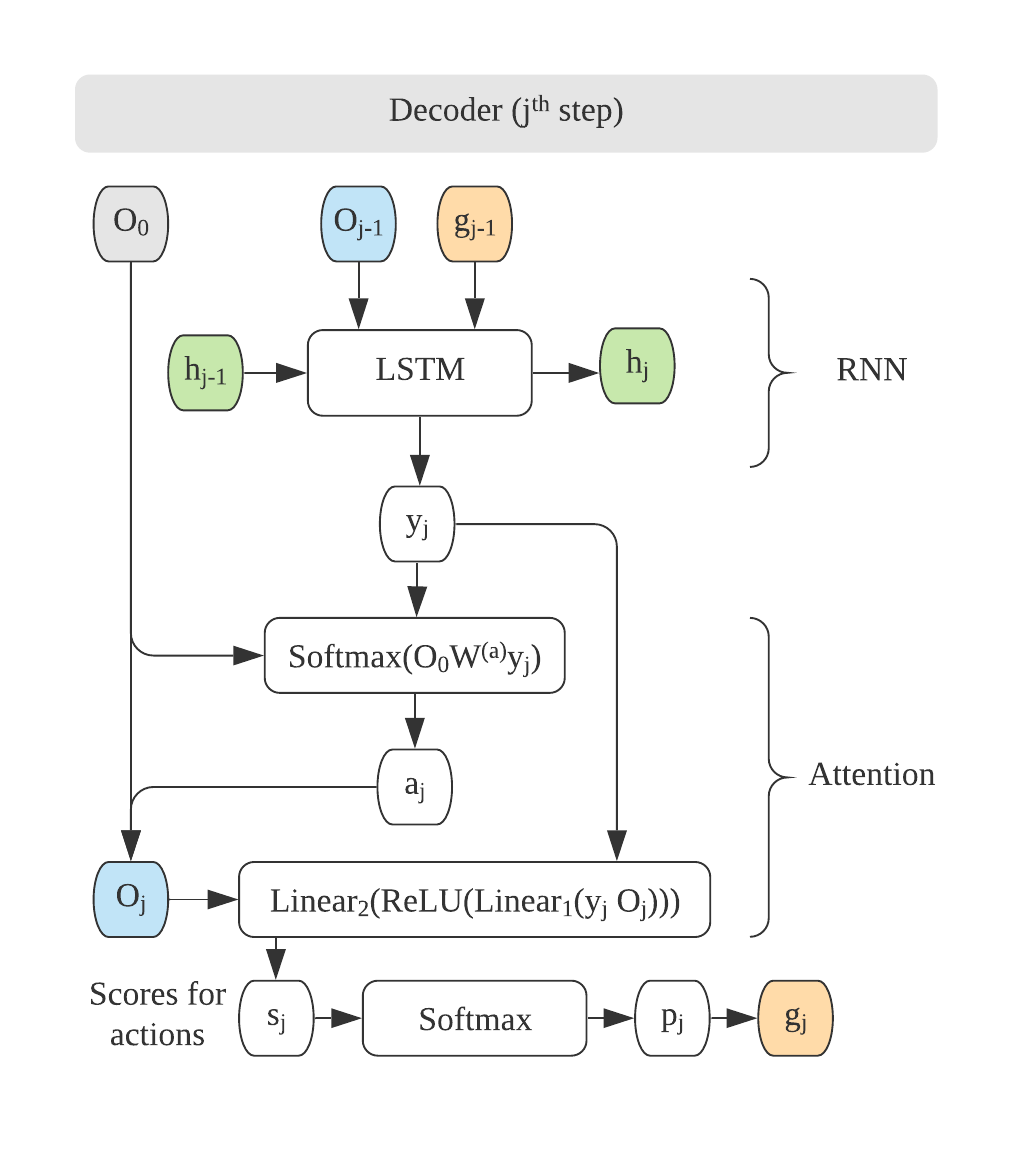}
    \caption{The architecture of the SMARTER decoder in the j\textsuperscript{th} step}
    \label{fig:arch_decoder}
\end{figure}

SMARTER's decoder, as shown in Figure \ref{fig:arch_decoder}, is based on the Transition-Based Parser architecture. It utilizes a Long Short-Term Memory (LSTM) layer with attention that selects parsing actions from a grammar over well-typed logical form. As described in the Language section, we have defined a grammar that specifies the allowed transitions on each intermediate parsing states. The decoder can choose which predicates to apply on the current state based on the distribution over grammar actions using an attention mechanism over the encoded question tokens in every step. When it arrives to the limit of the length of the logical form or reach the terminate state, the final logical form will serve us the output of the model. The execution of the logical form is a interval and all the relations to the events in the interval are our final outputs. \\

\section{Experiments}
\subsection{Experimental Settings}
\subsubsection{Datasets}
We use the datasets provided by TempEval-3 \cite{uzzaman-etal-2013-semeval}, the representative task in TIE, to train and evaluate our model,  which  are primarily English news with TimeML annotation. The training dataset consists of TimeBank, AQUAINT2, and TempEval-3 Silver, and the testing dataset is TempEval-3 Eval. We spilt 10\% of the training data for validation.

\subsubsection{Evaluation}
We extract the temporal relations from the datasets as the ground truth and calculate the recall rate of our annotated relations given the events, which accords with "Task-3 Relation Only" in TempEval-3.

\section{Conclusion}
We proposed SMARTER, an neural semantic parsing architecture for Temporal Information Extraction that includes an Event Annotator and a Relation Extractor. We adapted neural semantic parsing methods to the Temporal Reasoning domain to allow the parser extract temporal relations between events and time expressions in a paragraph effectively, outperforming other methods previously proposed. We also optimized the DPD training step for weak supervision dataset so that correct logical forms can be found efficiently. The BERT model is used in the Event Annotator and encoder, and they are proved to perform decently in our experiments.
\appendix

\bibliographystyle{named}
\bibliography{AAI_report}

\end{document}